\documentclass[conference]{IEEEtran}
\IEEEoverridecommandlockouts
\usepackage{cite}
\usepackage{amsmath,amssymb,amsfonts}
\usepackage{algorithmic}
\usepackage{graphicx}
\usepackage{textcomp}
\usepackage{xcolor}
\def\BibTeX{{\rm B\kern-.05em{\sc i\kern-.025em b}\kern-.08em
    T\kern-.1667em\lower.7ex\hbox{E}\kern-.125emX}}
\begin{document}

\title{A Transformer-based Multimodal Fusion Model for Efficient Crowd Counting Using Visual and Wireless Signals\\
}

\author{\IEEEauthorblockN{Zhe~Cui\IEEEauthorrefmark{1}, Yuli~Li\IEEEauthorrefmark{2},        and~Le-Nam~Tran\IEEEauthorrefmark{1}} \IEEEauthorblockA{\IEEEauthorrefmark{1}School of Electrical and Electronic Engineering, University College
Dublin, Ireland\\
\IEEEauthorrefmark{2} College of Electrical Engineering and Automation, Shandong University of Science and Technology, China\\
 Email: zhe.cui@ucdconnect.ie, 202283080050@sdust.edu.cn, nam.tran@ucd.ie} }

\maketitle

\begin{abstract}
Current crowd-counting models often rely on single-modal inputs, such as visual images or wireless signal data, which can result in significant information loss and suboptimal recognition performance. To address these shortcomings, we propose TransFusion, a novel multimodal fusion-based crowd-counting model that integrates Channel State Information (CSI) with  image data. By leveraging the powerful capabilities of Transformer networks, TransFusion effectively combines these two distinct data modalities, enabling the capture of comprehensive global contextual information that is critical for accurate crowd estimation. However, while transformers are well capable of capturing global features, they potentially fail to identify finer-grained, local details essential for precise crowd counting. To mitigate this, we incorporate Convolutional Neural Networks (CNNs) into the model architecture, enhancing its ability to extract detailed local features that complement the global context provided by the Transformer. Extensive experimental evaluations demonstrate that TransFusion achieves high accuracy with minimal counting errors while maintaining superior efficiency.
\end{abstract}


\section{Introduction}

\IEEEPARstart{A}{ccurately} estimating crowd density in diverse environments is a critical challenge with numerous applications for urban planning, security, and event management. Traditional methods, such as utilizing sensor data from smartphones, GPS, and Bluetooth \cite{franke2015smart}, aim to capture three-dimensional crowd information but often require specialized hardware, which can limit their applicability in real-world scenarios.

In recent years, computer vision-based technologies have advanced considerably, especially for human activity recognition  \cite{kong2022human}. These technologies analyze image and video data to recognize human postures, movements, and facial expressions with high accuracy and in real-time. Early vision-based methods relied on manually designed features for human sensing, which were later replaced by deep learning approaches that have significantly improved performance in tasks like classification, object detection, and segmentation \cite{ji2019survey}. Vision-based crowd counting methods, particularly those employing deep neural networks \cite{shi2018crowd,zhang2015cross,zhang2016single,babu2017switching,hossain2019crowd,jiang2020attention,lin2022boosting}, have demonstrated high accuracy. However, these methods still face challenges such as significant occlusions and poor lighting conditions, which can adversely affect crowd estimation accuracy.

Parallel to vision-based methods, WiFi-based human sensing has emerged as a promising alternative. These systems leverage changes in WiFi signals caused by human movement to infer activities \cite{zhou2023metafi++}. Advantages of WiFi-based systems include device-free operation, low cost, broad coverage, independence from lighting conditions, and privacy preservation \cite{zheng2019zero}. By modifying the device drivers of WiFi NICs like the Intel WiFi Link 5300, Channel State Information (CSI) can be extracted from the physical layer \cite{yousefi2017survey,yang2022efficientfi}. CSI provides detailed multi-channel data, enhancing the precision of WiFi-based sensing. This capability has led to increased interest in using WiFi signals from nearby devices for crowd sensing \cite{zou2018device}. Noteworthy developments include Domenico et al. \cite{di2016trained}, who used differential WiFi CSI with normalized Euclidean distance, and Chen et al. \cite{chen2017tr}, who relied on respiratory signals, though this method required individuals to remain still. Cheng et al. \cite{cheng2017device} achieved 88\% accuracy with a DNN-based system for counting up to 9 people, while Zou et al. \cite{zou2018device} developed WiFree with 92.8\% accuracy. Liu et al. \cite{liu2019deepcount} combined CNNs and LSTMs for 90\% accuracy, and Zhou et al. \cite{zhou2020device} used CSI with DNNs for high accuracy across different crowd sizes. Hou et al. \cite{hou2022dasecount} introduced DASECount, demonstrating over 92\% accuracy across various domains, and Choi et al. \cite{choi2022wi} developed Wi-CaL for localization and crowd counting with high accuracy in various room sizes. These methods leverage unique wireless signal patterns to detect crowd presence and movement, showing resilience in non-line-of-sight conditions. However, WiFi signals are susceptible to environmental influences.

To overcome the drawbacks of single-modal approaches, we consider the fusion of computer vision (CV) and WiFi-based sensing to improve crowd-counting accuracy. By integrating these modalities properly, the proposed approach aims to exploit their complementary strengths and address the weaknesses inherent in each method individually.
The main contributions of this paper are as follows: 
\begin{itemize} 
\item We introduce a novel crowd-counting model, TransFusion, which integrates Channel State Information and image data. The idea is to mitigate information loss and enhance crowd estimation accuracy by combining complementary features from both sources. 
\item TransFusion combines Convolutional Neural Networks with Transformer networks, addressing the limitations of Transformers in capturing fine-grained local features and enabling the extraction of both global and local characteristics. 
\item We provide extensive experimentation and evaluation of the proposed TransFusion model to demonstrate its effectiveness in achieving minimal error rates and maintaining high efficiency in crowd counting tasks. \end{itemize}

The remainder of this paper is organized as follows: Section \ref{model} describes the proposed TransFusion model. Section \ref{experiment} details the experimental setup and results. The paper concludes with Section \ref{conclusion}.

\enlargethispage{-1\baselineskip}
\addtolength{\topmargin}{0.05in}
\section{Methodology}
\label{model}

\subsection{System Overview}
\label{System Overview}

\begin{figure}[!t]
\centering
\includegraphics[width=0.9\linewidth]{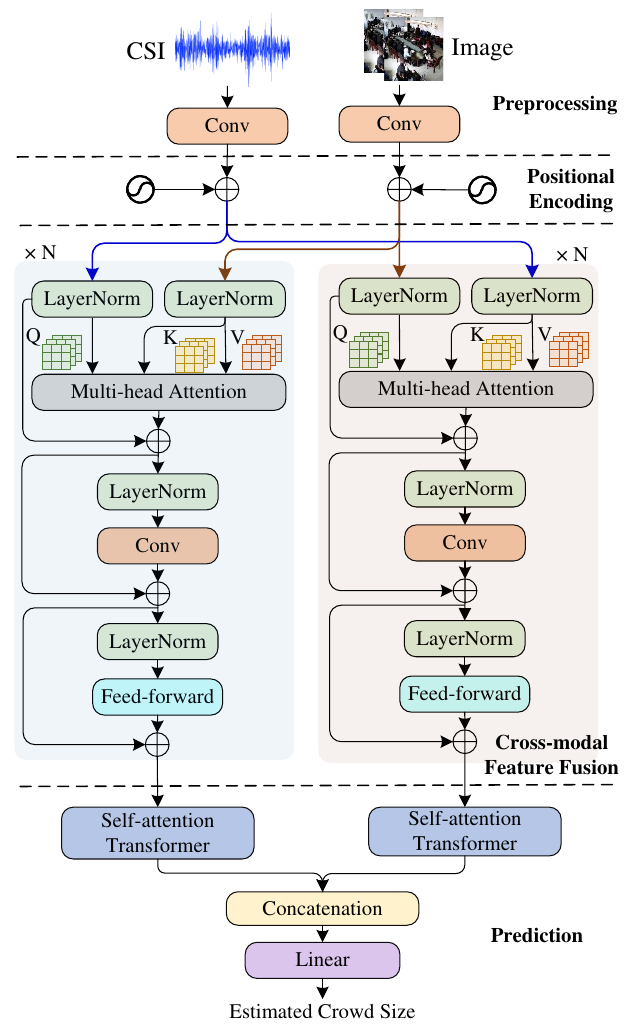}
\caption{The framework of the proposed TransFusion}
\label{structure}
\end{figure}

In our crowd counting prediction system, we consider two modalities: visual images (V) and Channel State Information measurements of WiFi (W). We denote the collected 2D visual images as $X^V\in \mathbb{R}^{h\times w\times c}$ (where $h$ and $w$ are the height and width of the image, and $c$ is the number of channels), and the amplitude of CSI measurements sequences as $X^{W} \in \mathbb{R}^{l^{W}\times d^{W}}$, respectively.\footnote{In our work we ignore the phase of CSI which is not reliable for the chosen WiFi NIC.} Here, $l^{W}$ and $d^{W}$ are the sequence length and feature dimension, respectively.
Our objective is to merge the multimodal measurements $X^{V}$ and $X^{W}$ via \textit{attention mechanism} across the two modalities. 

The structure of the proposed TransFusion is drawn in Fig. \ref{structure}. As can be seen clearly,  the proposed TransFusion comprises four key components: data preprocessing, positional encoding, cross-modal feature fusion, and prediction. 
To be specific, the TransFusion workflow initiates with data preprocessing, which involves denoising CSI measurements and transforming images. Following this, we apply positional encoding to both modalities. The encoded modalities are then fed to cross-modal attention for effective feature fusion. Finally, we leverage a linear attention transformer to collect temporal information, and make predictions. 
The details of TransFusion are described in the following subsections.

\enlargethispage{-1\baselineskip}
\subsection{Data Preprocessing}
\addtolength{\topmargin}{0.02in}
\subsubsection{Preprocessing} 
\
\newline
\indent CSI Denoising: Due to the susceptibility of CSI measurements to environmental influences, the data often contains noise that can impact accuracy. To enhance the precision of amplitude features, we employ the Hampel filter for denoising.

The Hampel filter operates as follows: Given a sequence of received amplitude CSI values denoted as $X^W_r=\left\{x^W_{r,1},x^W_{r,2},\dots x^W_{r,n}\right\}$, we initiate the denoising process by creating observation windows for each element with a width of $2K+1$. Subsequently, for each element in the sequence, if the difference between the sample and the median surpasses three times the standard deviation, we replace the sample with the median. The denoised CSI measurements are denoted as $X^W=\left\{x^W_{1},x^W_{2},\dots, x^W_{n}\right\}$. This process can effectively mitigate the impact of outliers and noise, contributing to more accurate amplitude measurements.

Visual images preprocessing: We divide the received 2D image $X^V\in \mathbb{R}^{h\times w\times c}$ into $p\times p$ blocks, resulting in a series of image blocks denoted as $X^{V}_{p} \in \mathbb{R}^{l^{V}\times d^{V}}$, where $l^{V}=\frac{h\times w}{p^{2}}$ and $d^{V}=p^2 \times c$. In our paper, we set $p$ to 16.

\subsubsection{Temporal Convolution Layer}
\
\newline
\indent The input CSI sequences inherently contain temporal information, thus, it's crucial to ensure that each element in the input sequences is contextually aware of its neighboring elements. To achieve this, we employ a temporal convolutional layer, denoted as $\bar{X}^W = \text{conv}(X^W_r, k^W)$, where $k^W$ denotes the kernel  for the WiFi modality. Similarly, for processed image sequences $X^{V}_{p}$ captured over the same time duration as WiFi signals, we utilize $\bar{X}^V = \text{conv}(X^V_{p}, k^V)$, where $k^V$ is the kernel for the image modality.

\subsection{Position Encoding}
To include the order information inherent in the received sequences, we add positional encoding to the sequences. Following \cite{vaswani2017attention}, we use sine and cosine functions of different frequencies:

\begin{subequations}
\begin{align}
PE(pos, 2j) &= \sin\left(\frac{pos}{10000^\frac{2j}{d}}\right)\label{pe_1} \\
PE(pos, 2j + 1) &= \cos\left(\frac{pos}{10000^\frac{2j}{d}}\right)\label{pe_2}
\end{align}
\end{subequations}
where $pos$ denotes the position, and $j$ represents the dimension. Thus, each dimension feature (i.e., column) of $PE$ encodes positional values in a sinusoidal pattern.

We add the positional encoding $PE$ to $\bar{X}^W$ and $\bar{X}^V$ to effectively encode positional information for each element at every time step:

\begin{equation}
Z^{{W,V}}_0 = \bar{X}^{{W,V}} + PE(l^{{W,V}}, d^{{W,V}})
\label{pr}
\end{equation}
where $Z^{{W,V}}_0$ denotes the respective low-level position-aware features for both modalities $W$ and $V$.
\addtolength{\topmargin}{0.01in}
\enlargethispage{-1\baselineskip}
\subsection{Cross-Modal Feature Fusion with Linear Attention}
In this section, we leverage attention-based cross-modal fusion to facilitate correlation between WiFi and image low-level position-aware features. Inspired by the attention mechanism, our proposed cross-modal feature fusion module introduces a latent attention-based adaptation across modalities—specifically, from WiFi to images ($W \rightarrow V$) and vice versa ($V \rightarrow W$). 

As illustrated in Fig. \ref{structure}, the proposed TransFusion architecture comprises two distinct streams of cross-modal models. Each stream is composed of $N$ layers of cross-modal attention blocks. Within each layer, there are sequentially stacked sub-layers, including multi-head self-attention, multi-scale CNN with adaptive scales, and feed-forward layers.

\subsubsection{Multi-Head Linear Attention}
\
\newline
\indent We employ a multi-head linear attention mechanism to facilitate the exchange of information between modalities.
Here, the sequence of one modality (referred to as the source modality) serves as the Query input, determining the length of the output sequence, while the sequences of the other modality (the target modality) are utilized as the Key and Value inputs.
The multi-head self-attention mechanism enhances the target modality by emphasizing relevant low-level features from the source modality using attention mechanisms. 
This mechanism incorporates multiple projection heads within a single attention layer, thereby enabling more comprehensive feature extraction through attention. 

Taking the example of passing wireless (W) information to vision (V), denoted as ``$W \rightarrow V$," it is computed feed-forwardly for $i=1,...,N$ layers as:
\begin{equation}
\label{cross_1}
\begin{split}
    &Z^{W\to V}_{0}  = Z^{V}_{0}\\
    &\hat Z^{W\to V}_{i} = \text{CF}^{W\to V}_{mul,i}(\text{LN}(Z^{W\to V}_{i-1}),\text{LN}(Z^{V}_{0}))+\text{LN}(Z^{W\to V}_{i-1})
\end{split}
\end{equation}
where $\text{CF}^{W\to V}_{mul,i}$ denotes the multi-head cross-modal attention at the $i$-th layer, and $\text{LN}$ signifies the normalization layer. Throughout this process, each modality continually updates its sequence with external information from the lower-level multi-head cross-modal attention module. Within the cross-modal attention block, low-level signals from the source modality are transformed into a set of distinct $K$ and $V$ values, facilitating interaction with the target modality.

We define the Querys as $Q^V=X^{V}I_q^W$, Keys as $K^W=X^{W}I_k^W$, and Values as $V^W = X^{W}I_v^W$, where $I_q^V$, $I_k^W$, and $I_v^W$ are weights. The latent adaptation from $W$ to $V$ is presented as the cross-modal attention $\text{CF}^{W\to V}(X^W, X^V)$:
\begin{equation}
\label{cross_2}
\text{CF}^{W\to V}(X^W, X^V) = softmax(Q^V {(K^W)}^{T})V^W
\end{equation}
where $I_q^W \in \mathbb{R}^{d^V \times d_k}$, $I_k^W\in \mathbb{R}^{d^W \times d_k}$, and $I_v^W\in \mathbb{R}^{d^W \times d_v}$;
$Q^V \in \mathbb{R}^{l^V \times d_k}$, $K^W \in \mathbb{R}^{l^W \times d_k}$, and $V^W \in \mathbb{R}^{l^W \times d_v}$. 
\enlargethispage{-1\baselineskip}
\subsubsection{Multi-scale CNN}
\
\newline
\indent While the above attention mechanism excels in capturing global features, it is less effective at extracting fine-grained local feature patterns from WiFi signals and visual images. Conversely, CNNs are highly effective at capturing local features but require numerous layers to extract global information. To address these limitations and effectively capture both local and global features from the input WiFi and image sequences, we employ a multi-scale CNN. This approach leverages different scale kernel sizes in a feed-forward manner:

\begin{equation}
\label{cross_convs}
Z^{W\to V}_{i} = f_{\theta,i}^{W\to V}(\text{LN}(\hat Z^{W\to V}_{i})+\text{LN}(\hat Z^{W\to V}_{i})
\end{equation}
where $f_{\theta}$ represents the multi-scale CNN parameterized by $\theta$. 
The multi-scale CNN processes the input sequences using kernels of varying sizes, allowing it to effectively capture both fine-grained local features and broader contextual information. This complementary approach ensures a comprehensive representation of the data, enhancing the model's ability to fuse information from both modalities for robust crowd-counting predictions.

\subsubsection{The Feed-forward module}
\
\newline
\indent Following the multi-scale CNN, the outputs are fed into a feed-forward module to facilitate non-linear transformations, enhancing the model's representational capacity. This module consists of fully connected layers with the Rectified Linear Unit (ReLU) activation function. The specific implementation is given by Equation \ref{ffn}:

\begin{equation}
FFN(x) = \max(0, x \cdot W_1 + b_1)W_2 + b_2
\label{ffn}
\end{equation}
Each layer of the Feedforward Neural Network is connected to a residual module, and the data is normalized through Layer Normalization.

\subsubsection{Self-Attention Module}

\
\newline
\indent Finally, we employ a Self-Attention Transformer for each stream to capture temporal information from the output features of each cross-modal transformer model. 

In this section, the self-attention module is primarily used to extract the final element of the sequence model and make predictions through fully connected layers.
The self-attention mechanism operates by automatically determining the allocation of weights to input items based on the inter-correlation information among input data features.
It decides the weight assigned to each input item by considering the interaction among input items. Unlike traditional attention mechanisms, the self-attention mechanism exhibits less dependency on external input information. This characteristic provides a significant advantage in capturing internal correlations within data or features and extracting deep-level features.
The self-attention mechanism allows the model to weigh the importance of different parts of the input sequence dynamically. By doing so, it enhances the model's ability to understand temporal dependencies and relationships within the data in each stream, leading to more accurate and robust predictions for crowd counting.

\enlargethispage{-1\baselineskip}
\subsubsection{Loss Function}
\
\newline
\indent Compared to other loss functions, L1 loss is less sensitive to outliers, providing a better balance for accurate and robust crowd counting predictions. Therefore, the loss function $\mathcal{L}$ is a standard L1 loss, defined as the Mean Absolute Error (MAE) between the ground truth label and the predicted value:
\begin{equation}
\mathcal{L} = \frac{1}{m} \sum_{i=1}^{m} |Y_i - \mathcal{F}(X^V_i,X^W_i)|
\label{L1}
\end{equation}
where $Y_i$ represents the ground-truth label corresponding to the $i$-th sample, and $\mathcal{F}(X^V_i, X^W_i)$ represents the predicted label produced by the proposed fusion model $\mathcal{F}()$.

\section{Experiments and evaluations}
\label{experiment}

\subsection{Experimental settings}
\subsubsection{Data Collection}

\begin{figure}[!t]
\centering
\includegraphics[width=1\linewidth]{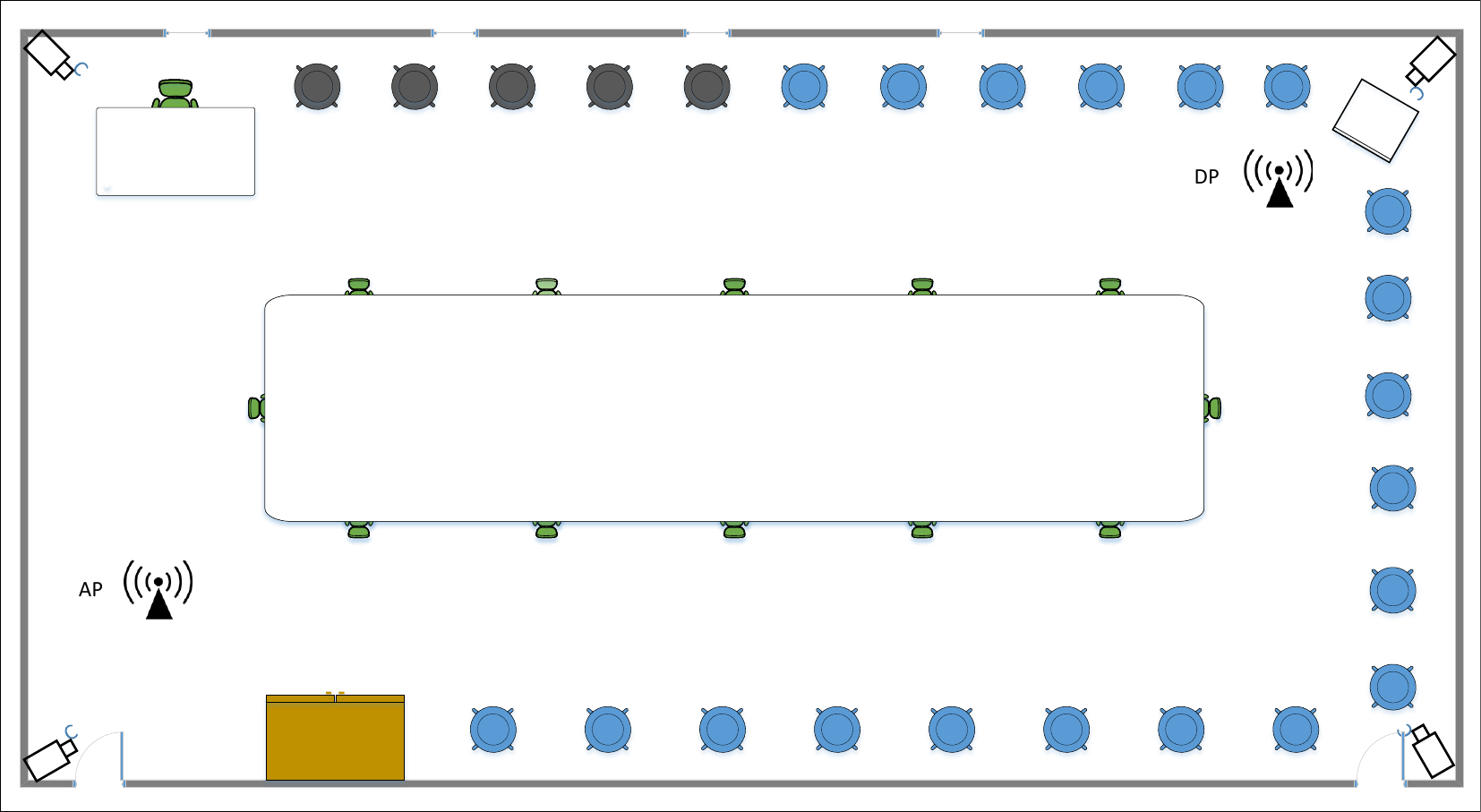}
\caption{The layout of experimental environment}
\label{meeting_room}
\end{figure}

\begin{figure}[!t]
\centering
\includegraphics[width=1\linewidth]{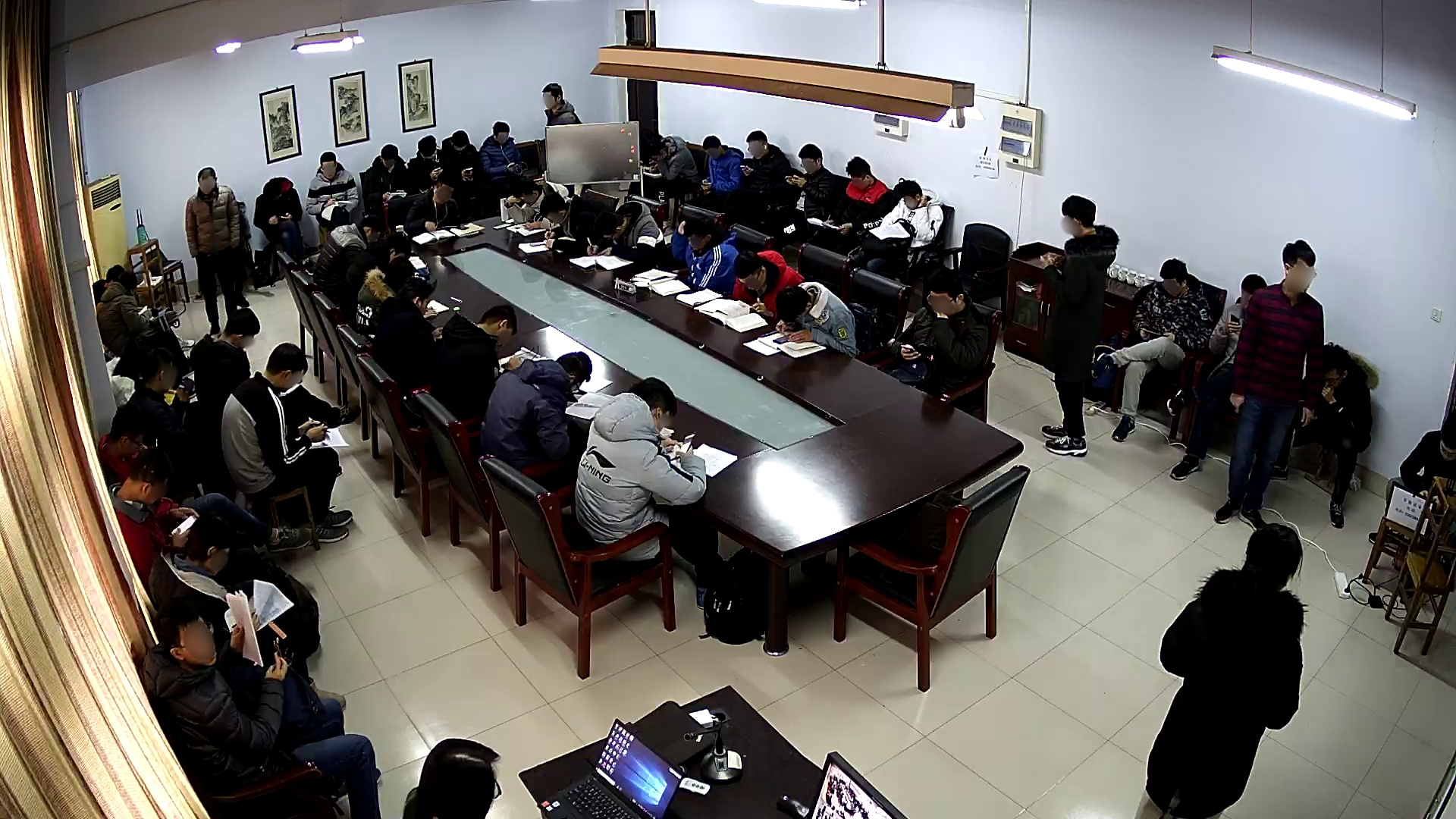}
\caption{State of people during experiment}
\label{people}
\end{figure}

\
\newline
\indent To evaluate the proposed model, we designed a prototype system to collect both WiFi signals and images for predicting crowd counts.
To record the WiFi signals, the prototype system employed two DELL laptops acting as the transmitter (TX) and the receiver (RX), both equipped with Intel 5300 network cards and the Linux 802.11n CSI Tool. The TX and RX were positioned on tripods at a height of 1.5 meters, with a separation of 10.5 meters between them. To ensure the collection of fine-grained information about crowd counting, the sampling rate was set to 500 Hz, and each sample collection time window was set to 4 seconds.  For different crowd counting scenarios, 100 collections were made for each scenario, with each collection capturing 2000 data packets. This resulted in a total of 200,000 data packets.
Simultaneously, to record the  image data, four Hikvision DS-2CC1185P cameras were installed at the four corners of the meeting room to record synchronized videos, ensuring real-time correspondence with the collected WiFi signals.

The experiments were conducted in a typical meeting room with an area of $10.8m\times7m$. The layout of the environment is depicted in Figure \ref{meeting_room}, featuring items common in office settings such as conference tables, chairs, air conditioning units, and storage cabinets. Figure \ref{people} illustrates the actual experimental setup. During data collection, a total of 44 volunteers actively participated, with the number of participants ranging from 0 to 44 in sequence. The volunteers were free to engage in various activities such as walking, talking, and working.

\enlargethispage{-1\baselineskip}
\subsubsection{Experimental Details}

\
\newline
\indent The model was implemented on a workstation equipped with an Intel(R) Xeon(R) Gold 5218R @ 2.10GHz twelve-core CPU and an RTX 3090 (24GB) GPU, using PyTorch 1.11.0 and Python 3.8. The dataset was randomly divided into training, validation, and testing sets with an 8:1:1 ratio. The batch size was set to 32, and the model was trained with an initial learning rate of $1 \times 10^{-3}$, running for a maximum of 200 epochs. The best-performing model on the validation set was selected for testing. The Adam optimizer was employed for training.

\subsubsection{Evaluation Metrics}
\
\newline
\indent Following common practice in crowd counting, we employ the mean absolute error (MAE), mean squared error (MSE), mean absolute percentage error (MAPE), and coefficient of determination ($R^2$) for evaluation. These metrics are defined as follows:
\begin{align}
    &\text{MAE} = \frac{1}{m} \sum_{i=1}^{m} |\mathcal{F}(X^V_i,X^W_i) - Y_i|\\
    &\text{MSE} = \frac{1}{m} \sum_{i=1}^{m} \left [\mathcal{F}(X^V_i,X^W_i) - Y_i)  \right ] ^2\\
    &\text{MAPE}  = \frac{1}{m} \sum_{i=1}^{m} \frac{\left|\mathcal{F}(X^V_i,X^W_i) - Y_i\right|}{Y_i}
\end{align}

\begin{equation}
R^2 = 1 - \frac{\sum_{i=1}^{m} \left [ \mathcal{F}(X^V_i,X^W_i) - Y_i \right ] ^{2} }{\sum_{i=1}^{m} \left [  \bar{Y}-Y_i\right ] ^2}
\label{L1}
\end{equation}
where $\bar{Y}=\frac{1}{m} \sum_{i=1}^{m}Y_i $ is the mean of the ground truth values.
MAE represents the average absolute difference between the predicted and actual counts, reflecting the model's accuracy. MSE measures the mean squared difference between the predicted and actual counts, indicating model robustness. MAPE calculates the average absolute percentage difference between the predicted and actual counts, highlighting model accuracy. 
$R^2$ assesses the proportion of the variance in the dependent variable that is predictable from the independent variables, with a higher value indicating better model fitting.

\subsubsection{Baselines}
\addtolength{\topmargin}{0.01in}
\enlargethispage{-1\baselineskip}
\
\newline
\indent To assess the effectiveness of the proposed method, we compare it with various crowd counting methods and fusion methods. The baselines include WiFi-based methods (FreeCount \cite{zou2017freecount}, WiCount \cite{liu2017wicount}) and multimodal fusion methods, including Early Fusion LSTM (EF-LSTM) \cite{niu2024follower}, Late Fusion LSTM (LF-LSTM) \cite{niu2024follower}, Low-rank Multimodal Fusion method (LMF) \cite{liu2018efficient}, and Transformer Routing (TRAR) \cite{zhou2021trar}.








\subsection{Results and Discussions}
\subsubsection{Performance Overview}
\
\newline
\indent Table \ref{compare} compares the performance of our proposed TransFusion model with various other models, including WiFi-based models (FreeCount and WiCount) and multimodal fusion models (EF-LSTM, LF-LSTM, LMF, and TRAR).

Among the WiFi-based models, FreeCount demonstrates the best performance with an MAE of 4.7952. This suggests that the deep neural network (DNN) in WiCount may be overfitting the data, resulting in poorer performance compared to traditional machine learning models like SVM when solely using WiFi signals.
For multimodal fusion models, LF-LSTM achieves the best performance with an MAE of 0.3812. LSTM networks are well-suited for extracting and training temporal information features, and previous studies have shown that late fusion outperforms early fusion. Late fusion methods can handle asynchronous data more effectively and are easily extensible based on the number of modes, allowing better modeling of the unique characteristics of each modality.
Compared to the previously best-performing LF-LSTM model, the proposed TransFusion model significantly reduces the MAE by 60.49\%. Similarly, it reduces the MSE by 49.37\% and MAPE by 35\%, while increasing the $R^2$ by 0.23\%. This demonstrates that the cross-modal fusion based on Transformer, the extraction of local features through convolutional neural networks, and the improvement of linear attention are effective. The proposed algorithm shows better generalization performance compared to other methods.
\begin{table}[!ht]
    \centering
    \caption{Performance comparison of different algorithms}
    \begin{tabular}{ccccc}
    \hline
        Model & MAE & MSE & MAPE & $R^{2}$ \\ \hline
       FreeCount \cite{zou2017freecount} & 4.7952 & 44.7958  & 0.2804 & 0.7381 \\ 
        WiCount \cite{liu2017wicount} & 5.8968 & 59.2196 & 0.3750 & 0.7005 \\ \hline 
        EF-LSTM \cite{niu2024follower}& 1.7488 & 13.7959 & 0.1467 & 0.6977 \\ 
        LF-LSTM \cite{niu2024follower} & 0.3812 & 0.3405 & 0.0180 & 0.9967 \\ 
        LMF  \cite{liu2018efficient}& 3.8853 & 33.9677 & 0.2741 & 0.8032 \\ 
        TRAR \cite{zhou2021trar}& 1.4865 & 12.6528 & 0.1348 & 0.7843 \\  \hline
        \textbf{TransFusion} & \textbf{0.2069} & \textbf{0.3831} & \textbf{0.0164} & \textbf{0.9978} \\ \hline
    \end{tabular}
    \label{compare}
\end{table}
\enlargethispage{-1\baselineskip}
\subsubsection{Ablation Study}
\
\newline
\indent We conducted an ablation study to evaluate the contributions of different components (Vision stream, WiFi stream, Multi-scale CNN, Linear attention) to the model performance. 
In each test, we ablated a specific component from the full TransFusion model.
The results are shown in Table~\ref{ablation}. 

\begin{table*}[!ht]
\centering
\caption{Ablation study results compared with the full TransFusion  model}
\begin{tabular}{l|c|c|c|c|c|c|c|c}
\hline
Model               & MAE & $\Delta$ & MSE & $\Delta$ & MAPE &  $\Delta$& $R^{2}$ &$\Delta$  \\\hline
TransFusion          
&  0.2069   & - &  0.3831   & - &  0.0164    & - &  0.9978 & - \\
~~-Vision stream    
&  2.0282   & +1.8213 &    13.6173  & +13.2342 & 0.1055 &  +0.0891    & 0.9247  & -0.0731 \\
~~-WiFi stream      
&  1.5856   & +1.3787 &  10.0073   & +9.6242 &  0.1367    &  +0.1203 & 0.9478  & -0.0500 \\
~~-Multi-scale CNN  
&   0.6676  & +0.4607 &  3.4731 & +3.0900  & 0.0384 & +0.0220&  0.9799 & -0.0179 \\
~~-Linear attention 
&  0.2069 &+0.0563 & 0.3831 & +0.2107& 0.0164 &+0.0047 & 0.9978 &-0.0012 \\ 
\hline
\end{tabular}
\label{ablation}
\end{table*}

From Table \ref{ablation}, we can observe the significant impacts of each ablated cpmponent on TransFusion's performance. Removing the vision stream led to the most drastic degradation, with increases in MAE by 1.8776 and MSE by 13.6173. Similarly, removing the WiFi stream also resulted in considerable degradation, showing an increase in MAE by 1.4350 and MSE by 10.0073. These findings suggest that models relying solely on CSI or image information individually exhibit poor performance.  This could be attributed to training exclusively on a single modality leading to information loss, which adversely affects prediction accuracy.
Removing the multi-scale CNN led to an increase in MAE (+0.5170), MSE (+3.3007), and a decrease in 
$R^2$ (-0.0191). This indicates that the Transformer alone is limited to extracting global features, neglecting local detailed features, whereas the multi-scale CNN can effectively capture local patterns across different scales.
Ablating the linear attention mechanism resulted in a slight increase in MAE, MSE, and MAPE, and a slight decrease in  $R^2$, suggesting that linear attention enhances the model's performance by helping to better integrate CSI and image modalities.
Overall, the proposed method effectively integrates CSI and image modalities, capturing more information and significantly improving prediction performance, demonstrating superior overall efficiency.

\section{Conclusion}
\label{conclusion}
In this paper, we propose TransFusion, a novel multimodal fusion model designed for accurate crowd counting using WiFi signals and visual images. Our approach leverages the strengths of Transformer networks for effective multimodal fusion while incorporating multi-scale convolutional neural networks to capture both local and global features from the input modalities. To address the high computational complexity often associated with Transformers, TransFusion employs linear attention mechanisms, significantly reducing model complexity and accelerating the training process. Extensive experiments conducted in a typical indoor environment demonstrate the superior effectiveness of TransFusion in enhancing both the accuracy and efficiency of crowd counting tasks.
\enlargethispage{-1\baselineskip}

\bibliographystyle{IEEEtran}
\bibliography{IEEEabrv.bib,ref.bib}


\end{document}